\documentclass[a4paper, 10pt, conference]{ieeeconf}     

\IEEEoverridecommandlockouts                              
                                                          
\usepackage{cite}
\usepackage{amsmath,amssymb,amsfonts}
\usepackage{algorithm}
\usepackage{algorithmic}
\usepackage{graphicx}
\usepackage{textcomp}
\usepackage{caption}
\usepackage{subcaption}
\usepackage{xcolor}
\usepackage{tabularx}
\usepackage{arydshln}
\usepackage{multirow}
\usepackage{amsmath}
\usepackage{cancel}
\usepackage{tikz}
\usetikzlibrary{matrix}
\usetikzlibrary{patterns}
\def\BibTeX{{\rm B\kern-.05em{\sc i\kern-.025em b}\kern-.08em
    T\kern-.1667em\lower.7ex\hbox{E}\kern-.125emX}}
\begin{document}

\title{Conference Paper Title*\\
{\footnotesize \textsuperscript{*}Note: Sub-titles are not captured in Xplore and
should not be used}
\thanks{Identify applicable funding agency here. If none, delete this.}
}

\title{Hybrid Force Motion Control with Estimated Surface Normal for Manufacturing Applications}
\author{Ehsan~Nasiri and Long~Wang
\thanks{E.~Nasiri and L.~Wang are with the Department
of Mechanical Engineering, Stevens Institute of Technology, New Jersey,
NJ, 07030, USA, e-mail: long.wang@stevens.edu}
\thanks{This research was supported in part by NSF Grant CMMI-2138896.}}

\maketitle

\begin{abstract}
This paper proposes a hybrid force-motion framework that utilizes real-time surface normal updates. The surface normal is estimated via a novel method that leverages force sensing measurements and velocity commands to compensate the friction bias.

This approach is critical for robust execution of precision force-controlled tasks in manufacturing, such as thermoplastic tape replacement that traces surfaces or paths on a workpiece subject to uncertainties deviated from the model.\par
We formulated the proposed method and implemented the framework in \textit{ROS2} environment. The approach was validated using kinematic simulations and a hardware platform.
Specifically, we demonstrated the approach on a 7-DoF manipulator equipped with a force/torque sensor at the end-effector. \\
\end{abstract}

\section{Introduction}
The utilization of robot force control across various sectors such as manufacturing, healthcare, and more is on the rise. 
On the other hand, force control tasks on a workpiece represent a complex challenge, requiring the robot to dynamically and autonomously adjust its path.
Such adaptation is essential to manage the uncertainties encountered in the workpiece during the manufacturing process \cite{Stolt2012,Yin2012}.
In numerous scenarios, robots are required to trace a contour or path while exerting a force perpendicular to the surface. 
This capability is particularly relevant in processes such as grinding \cite{Thomessen2000}, thermoplastic tape replacement \cite{Ahrens1998} and injection molding \cite{Ohba2009,Huang2009}. 

Many researchers have presented hybrid methods for the control of position and force in the manufacturing process, for instance \cite{Ferretti1997,Natale1999,Yoshikawa1993}.
Solanes et al. use the hybrid force-motion control of robots for surface polishing, employing a task priority method \cite{Solanes2018}.
In \cite{Namvar2005}, the authors proposed an adaptive hybrid force-motion control method to guarantee asymptotic tracking of desired motion-force trajectories in an unknown environment.
The application of a Kalman filter-based hybrid force-motion control to estimate the actual contact point has been addressed in \cite{Xia2016}. The development of an adaptive fuzzy control-based hybrid framework, aimed at improving the manipulator's performance in uncertain environments, has been presented in \cite{Wang2021}. One of our envisioned future applications is thermoplastic fiber layup for 3D surfaces of workpieces with complex geometry.

Precise hybrid motion-force control is greatly sought after in robotic manufacturing processes.
However, such precision is constrained and affected by environmental factors like friction. 
Thus, friction compensation becomes essential for the effectiveness of high-precision motion-force control systems.
Although there is a broad spectrum of methods for friction compensation in robotics, as documented in the literature, references such as \cite{Bona2005,Chen2009, Dupont94,Lee1995,Yang2018, Olsson1999, Misovec1999, Kermani2004, Avelar2016, Huang2016}, indicate that diverse friction models, essential for this purpose, have been explored (these methods can broadly be categorized into two main types: static models and dynamic models \cite{Hess1990, Canudas1995, Swevers2000, Lampaert2002}). Yet, there are very few studies specifically addressing friction compensation in hybrid force-motion control.
The challenges associated with friction compensation at joint-level  for hybrid force-motion control, have been the subject of research over the past years, as in \cite{ Jatta2006,Cao2019}.\par
\begin{figure}[t]
    \centering
    \includegraphics[width=1\columnwidth]{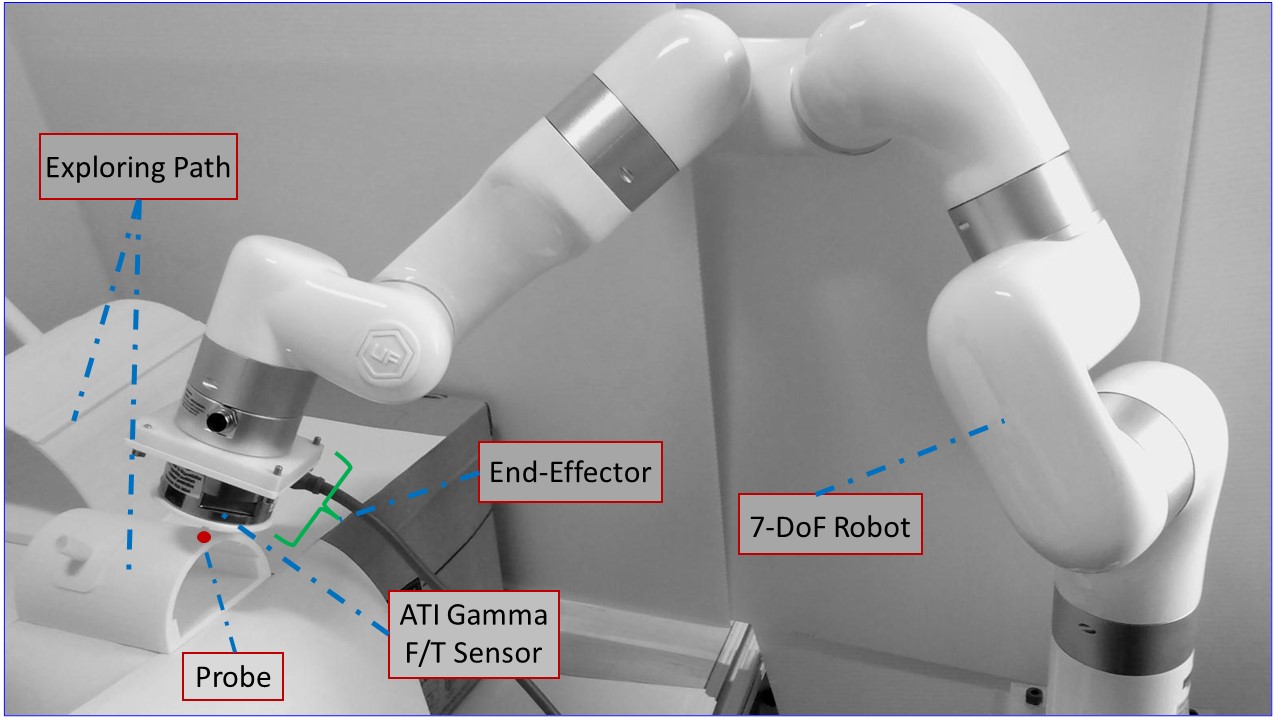}
   
    \caption{The Robotic System Framework: A 7-DoF Manipulator Equipped with an F/T Sensor and Probe at the End-Effector, Including Workpieces.}
   
    \label{system}
\end{figure} 
When a robot manipulator's end-effector comes into contact with an environment, there is a dynamic interaction between the manipulator and the environment. 
This interaction produces reaction forces that need to be managed effectively at the contact point.
Given the complexities inherent in advanced industrial processes, it is crucial to investigate surface normal estimation. 
Therefore, accurately estimating this norm, especially considering the friction on contact surfaces, is essential for precise manufacturing.
Failing to do so can lead to undesired outcomes, potentially causing the task to fail.
So far, a few studies have focused on motion-force control that includes estimating the surface normal during the trajectory\cite{Qian2019,Li2022}.\par
In this study, we employ a 7-DoF manipulator equipped with an F/T sensor and a probe at the end-effector.
This setup is used to execute tasks on various workpieces using the proposed hybrid force-motion control strategy. 
The robot utilizes a novel algorithm primarily to estimate the direction of the surface normal in real-time, as detailed in this study, while also determining the Coulomb coefficient of the surface. 
This method leverages force sensor readings to compensate for surface friction, thereby facilitating the estimation of the surface normal. 
The contributions of this work include:

\begin{itemize}
  \item A novel hybrid force-motion control framework is introduced, featuring a newly developed methodology to estimate surface normal.
  This method computes the surface normal by constructing a surface friction model. It does so by utilizing the force sensor feedback and estimating the surface Coulomb coefficient. 
  \item Validations were carried out in simulations and experiments, on a 7-DoF robot manipulator equipped with an F/T sensor.
\end{itemize}\par

\section{Modeling of the Robotic System}\label{system_modeling_section}
\subsection{Kinematic Modeling}
Throughout this paper, we adopt the nomenclature in Table\ref{nomenclature}. 
Our robotic system framework, as shown in Fig.\ref{system}, is developed within the \emph{\text{ROS2}} environment. 
The 7-DoF robotic system manipulates a force-sensing probe attached to the end-effector, comprising an F/T sensor.\par
\begin{table}
    \centering
    \caption{Nomenclature for Kinematics}
    \renewcommand{\arraystretch}{1.2} 
    \begin{tabular}{m{.15\columnwidth} m{.65\columnwidth}}
        \hline
        \textbf{Symbol} & \textbf{Description} \\
        \hline
        $\boldsymbol{{\emph{ee}}}$ & End-effector.  \\
        \hdashline
        $\boldsymbol{{\mathbf{x}_{\text{ee}}}}$ & Probe's center position.  \\
        \hdashline
        $\boldsymbol{J}_{{ee}}$ & Jacobian of the end-effector (\(\emph{ee}\)).  \\
        \hdashline
        $\boldsymbol{e}_{{ee}}$ & Probe (end-effector) error. \\
        \hdashline
        $\boldsymbol{K}_{{ee}}$  & Proportional $3\times3$ diagonal gain. \\
         \hdashline
        $\boldsymbol{K}_{\text{adm}}$  & Admittance gain.  \\
        \hdashline
        $\boldsymbol{K}_{m}$, $\boldsymbol{K}_{f}$ &  Diagonal $3\times3$ gains in hybrid force-motion control command. \\
        \hdashline
        $\boldsymbol{K}_{m\perp}$, $\boldsymbol{K}_{f\perp}$ &  Projected positional and derivative, diagonal gains in hybrid force-motion control command. \\
        \hdashline
        $\boldsymbol{N},\boldsymbol{T}$ & Desired force and motion control directions.   \\
         \hdashline
        $\boldsymbol{\hat{n}_{\text{surf}}}$, $\boldsymbol{{{n}}_{\text{ee}}}$ & Estimation of surface normal and end-effector directions.   \\
         \hdashline
        $\boldsymbol{\gamma}$ & The angle between the estimated surface normal and the estimated probe directions.   \\
        \hdashline
         $\boldsymbol{\mu}$ & The Coulomb coefficient of friction.   \\
        \hdashline
         $\boldsymbol{f}_{\text{des}}$ & The reference desired force.   \\
        \hdashline
        $\boldsymbol{f}_{\tau}$ & The surface friction force.   \\
        \hdashline
        $\boldsymbol{f}_{s}$ & Force sensor data.   \\
        \hdashline
        $\boldsymbol{f}_{\perp}$ & The projected surface friction force onto "null space" of robot velocity.   \\
        \hdashline
        $\boldsymbol{f}_{v}$ & The projected surface friction onto robot velocity.   \\
        \hdashline
         $\boldsymbol{{\Omega}}_m$, $\boldsymbol{{\Omega}}_f$ &  Motion and force projection matrices.  \\
         \hdashline
        $\boldsymbol{{\rho}}$ & A criterion that could be chosen to maximize an objective function \(g\), within the scope of resolving the robot's redundancy.\\
        \hdashline
        $\boldsymbol{\dot{\Phi}}$ & The configuration space velocity.   \\
        \hline
    \end{tabular}
    \label{nomenclature}
\end{table}
The velocity of the robot's end-effector, denoted as $\mathbf{\dot{x}}_{\text{ee}}$, is given by,
\begin{equation}
   \mathbf{\dot{x}}_{\text{ee}}= \mathbf{J}_{\text{ee}}(\mathbf{{\Phi}}) \;\dot{\mathbf{{\Phi}} }
    \label{ins_velocity2}
\end{equation}
Where, \(\mathbf{J}_{\text{ee}}\) represents the manipulator's Jacobian matrix which maps joint velocities to the end-effector's velocity.
The general solution that satisfies~\eqref{ins_velocity2}, is as follow:
\begin{align}
         \mathbf{\dot{\Phi}} _{\text{cmd}} = \mathbf{J}_{\text {ee}}^{\dagger} \mathbf{\dot{x}}_{\text{ee}} + \left(\mathbf{I}-\mathbf{J}_{\text {ee}}^{\dagger} \mathbf{J}_{\text {ee}}\right) \boldsymbol{\rho}
    \label{redundacy}
\end{align}
Here, we substitute the end-effector velocity \(\mathbf{\dot{x}}_{\text{ee}}\), as follows,
\begin{align}
    & \dot{\mathbf{x}}_{\text{ee}} =  \mathbf{K}_{\text{ee}} \mathbf{e}_{\text{ee}}, \quad \mathbf{K}_{\text{ee}} \in    \mathbb{R}^{3\times3}
    \label{new_command} \\[2pt]
    & \mathbf{e}_{\text{ee}} = 
    \mathbf{x}_{\text{des}} - \mathbf{x}_{\text{ee}} 
    \label{error_tot}
\end{align}
\(\boldsymbol{\rho}\) is a residue vector that can be selected to resolve the redundancy of the robot. In~\eqref{redundacy} the error, denoted as \( \mathbf{e}_{\text{ee}} \), is defined as the distance between the probe (\text{\emph{ee}}) and the targeted contact surface. \par
Considering the position of the contact point \(\mathbf{x}_\text{cnt}\), it is defined as follows:
\begin{equation}
    \mathbf{x}_{\text{cnt}}= \mathbf{x}_{\text{ee}} + {\mathbf{{n}}_{\text{surf}}} \hspace{2pt} {d}, \quad  d \in \mathbb{R}
    \label{extra_constraint}
\end{equation}
An estimation form of equation~\eqref{extra_constraint} can be given by,
\begin{equation}
    \mathbf{\hat{x}}_{\text{cnt}}= \mathbf{x}_{\text{ee}} + {\mathbf{\hat{n}}_{\text{surf}}} \hspace{2pt} {d} 
\end{equation}
Here, ${\emph{\text{d}}}$ signifies a scalar offset. We managed to maintain an offset distance between the probes and the goal, for instance, a 5mm offset between the center point of the probe and the desired contact point along the z-axis. Then,~\eqref{error_tot} can be rewritten,
\begin{align}
    \mathbf{e}_{\text{ee}}=\mathbf{x}_{\text{cnt,des}} - \mathbf{\hat{x}}_{\text{cnt}} 
    \label{new_error}
\end{align}
Taking into account equations~\eqref{error_tot} to~\eqref{new_error} and considering the offset, the error is expressed as follows:
\begin{equation}
    \mathbf{e}_{\text{ee}} =  (\mathbf{x}_{\text{cnt,des}} - \mathbf{\hat{x}}_{\text{cnt}}) - {{\hat{\mathbf{n}}_{\text{surf}}} \hspace{2pt} {d}}
    \label{new_error_tot}
\end{equation}\par
In the subsequent sections, we introduce a method for real-time estimation of the surface normal, denoted as \(\mathbf{\hat{n}}_{\text{surf}}\). 
This method facilitates the redefinition of the end-effector error, as outlined in~\eqref{new_error_tot}, thereby transitioning to a more robust approach 
Ultimately, the simulation and experimental validations section will present a in-depth comparative evaluation, focusing on the accuracy of the proposed method.
This will be illustrated through detailed plots and figures.\par
\begin{figure}
    \centering
    \includegraphics[width=1\columnwidth]{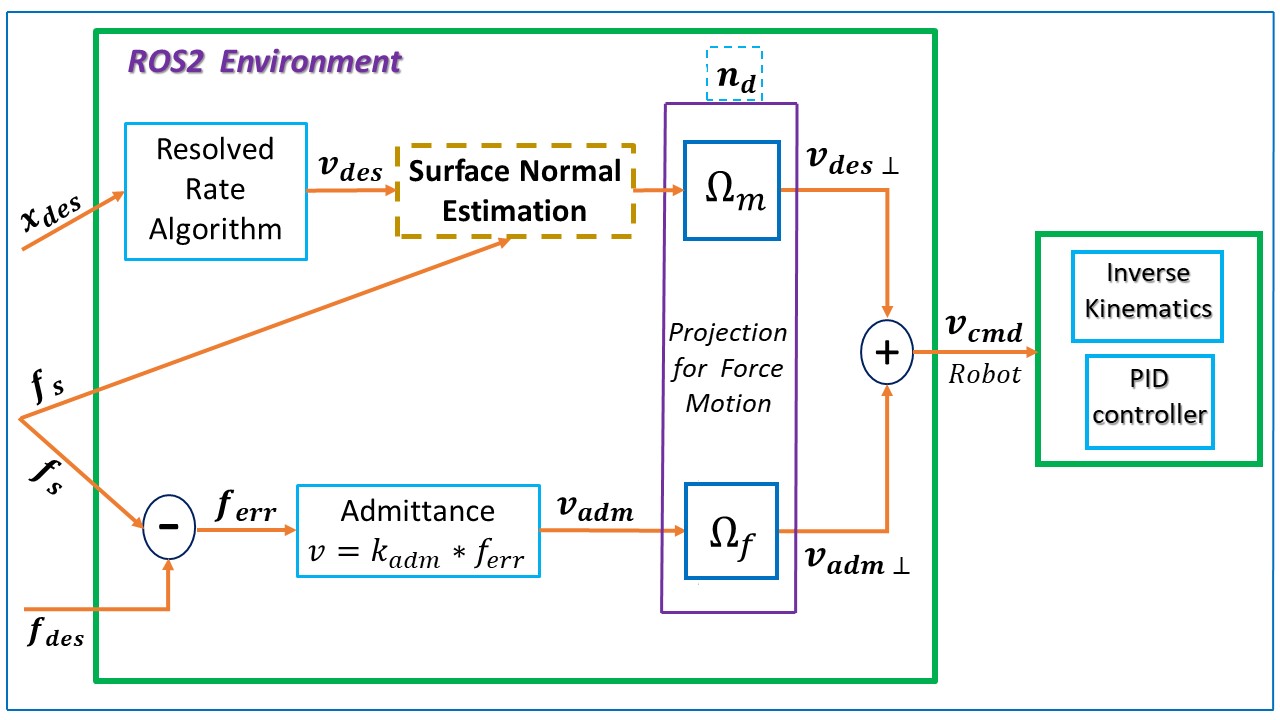}
    \vspace{-10pt}
\caption{Hybrid Force-Motion Controller}
    \label{controller}
    \vspace{-10pt}
\end{figure}
\subsection{Hybrid Force-Motion Control}
To facilitate hybrid force-motion, a dedicated node was developed within the ROS2 environment.
This integration is schematically depicted in Fig.\ref{controller}. 
The mechanism for motion-force projection, derived by inspiration from the research conducted by\cite{Khatib1987} and\cite{Featherstone1999}.
\begin{align}
    \mathbf{\Omega}_f &= \mathbf{N}(\mathbf{N}^\mathrm{T} \mathbf{N})^{-1} \mathbf{N}^\mathrm{T} = \mathbf{I} - \mathbf{{\Omega}}_m  \label{projection1} \\
    \mathbf{{\Omega}}_m &= \mathbf{T}(\mathbf{T}^\mathrm{T} \mathbf{T})^{-1} \mathbf{T}^\mathrm{T} = \mathbf{I} - \mathbf{\Omega}_f
    \label{projection2}
\end{align}\par
In the context of manipulation, a velocity command, denoted by \( \mathbf{v_{\text{des}}}\), is formulated using the resolved rate algorithm, as described by Whitney\cite{Whitney1969}. 
The velocity command \(\mathbf{v_{\text{des}}}\), derived from the resolved rate control block, is subsequently input into the surface normal estimation block. 
Additionally, this block receives input from force sensor measurements.
These measurements are crucial in estimating both the surface normal and the surface friction coefficient.
The process of this estimation follows the methodology detailed in Algorithm\ref{table2}.\\
In addition, a separate velocity command, denoted as \( \mathbf{v_{\text{adm}}} \), is generated based on the force error, $\mathbf{f}_{\text{err}}$, and an admittance gain, \( \mathbf{K_{\text{adm}}} \).
\begin{align}
    &\mathbf{v}_{\text{adm}}= \mathbf{K}_{\text{adm}} \hspace{2pt} \mathbf{f}_{\text{err}}, \quad \mathbf{K}_{\text{adm} }  \in \mathbb{R}^{3\times3}\\
   &  \mathbf{f}_{\text{err}} =  \mathbf{f}_{\text{des}} - \mathbf{f}_{s}
\end{align}\par
As illustrated in Fig.\ref{controller}, the velocity vectors \( \mathbf{v_{\text{des}}} \) and \( \mathbf{v_{\text{adm}}} \) undergo projection onto their respective motion-force planes. 
These projections are facilitated by the force-motion projection matrices, as defined in~\eqref{projection1},\eqref{projection2}. 
This step results in the respective projected command velocities \( \mathbf{v}_{\text{des}\perp} \) and \( \mathbf{v}_{\text{adm}\perp} \). 
Following this, the combined velocity command \( \mathbf{v_{\text{cmd}}} \) is dispatched to the robot controller. 
This is executed through a ROS2 node, which publishes the requisite joint velocities.

It should be noticed that in~\eqref{projection1}, \(\mathbf{N}\) is defined to be the desired force control direction (denoted by $\mathbf{n}_d$),
\begin{equation}
    \mathbf{N} = \mathbf{n}_{d} = \begin{bmatrix} n_{x} & n_{y} & n_{z} \end{bmatrix}^{\mathrm{T} }
\end{equation}
The robot's inverse kinematics, along with the necessary PID controller for guiding the manipulator in hybrid force-motion control, are implemented in a manner similar to that described in~\eqref{redundacy}.\par
According to Fig.\ref{controller}, the resulting hybrid force-motion control command in the configuration space can be expressed as follows:
\begin{equation}
    \mathbf{\dot{\Phi}}_{\text{cmd}}= \mathbf{\dot{\Phi}}_m +\mathbf{\dot{\Phi}}_f
    \label{decoupled_hybrid}
\end{equation}
Here, \(\mathbf{\dot{\Phi}}_{\text{cmd}}\) represents the resultant integrated motion and force velocity control commands.
\begin{align}
&{\mathbf{\dot{\Phi}}}_{{m}}= 
      {\mathbf{J}_{\text{ee}}^\dagger} (\mathbf{\dot{{x}}}_{\text{des},m}+  \boldsymbol{{{\xi}}}_m) 
      \label{hyb_motion} \\
&{\mathbf{\dot{\Phi}}}_{{f}} = 
      {\mathbf{J}_{\text{ee}}^\dagger} (\mathbf{\dot{{x}}}_{\text{des},f} +  \boldsymbol{{{\xi}}}_f) 
      \label{hyb_force}
\end{align}
\(\mathbf{\dot{{x}}_{\text{des}}} \) and \(\boldsymbol{{{\xi}}} \), in~\eqref{hyb_motion},\eqref{hyb_force} are detailed as follows, referring to Fig.\ref{controller}:
\begin{align} 
& \boldsymbol{{\xi}}_m =\mathbf{K}_{m \perp} \hspace{1pt}{\mathbf{e}}_{m} \label{xi_1}\\
&\mathbf{\dot{{x}}}_{\text{des},m} = \mathbf{v}_{\text{des}\perp} \\[5pt]
    & \boldsymbol{{\xi}}_f =\mathbf{K}_{f \perp} \hspace{1pt} {\mathbf{e}}_{f} \label{xi_3} \\
    &\mathbf{\dot{{x}}}_{\text{des},f} = \mathbf{v}_{\text{adm}\perp}  \label{xi-4}
\end{align}
The projected gain \(\mathbf{K}_{m \perp}\) is given by, 
\begin{align}
   \mathbf{K}_{m \perp} = \mathbf{K}_{m} \left(\mathbf{I} - \boldsymbol{\Omega}_f\right), \quad \mathbf{K}_{m }  \in \mathbb{R}^{3\times3}
   \label{motion_projected_gain}
\end{align}
Similar to~\eqref{motion_projected_gain}, for the force velocity command~\eqref{hyb_force}, the gain is given as follows:
\begin{align}
    \mathbf{K}_{f \perp} = \mathbf{K}_{f} \hspace{1pt}  \boldsymbol{\Omega}_f, \quad \mathbf{K}_{f } \in \mathbb{R}^{3\times3} \label{force_projected_gain} 
\end{align}
Using equations~\eqref{motion_projected_gain} and~\eqref{force_projected_gain}, the motion-force errors, which are expressed as follows, can be projected onto the motion-force directions, respectively:
\begin{align}
  {\mathbf{e}}_{m},{\mathbf{e}}_{f}=  (\mathbf{x}_{\text{des}} - \mathbf{x}_{\text{ee}}) -  \hspace{2pt} \mathbf{{d}}_h,  \quad \mathbf{d}_{h}  \in \mathbb{R}^{3\times1}\label{motion_error}
\end{align}
\(\mathbf{{d}}_h\) represent the offsets in the motion-force directions. \par
The term \(\mathbf{v}_{\text{cmd}}\), as illustrated in Fig.\ref{controller}, represents the integrated command velocity derived from the decoupled inputs of the force-motion control components. This velocity is defined as follows:
\begin{equation}
     \mathbf{v}_{\text{cmd}}= \mathbf{v}_{\text{adm}\perp} +\mathbf{v}_{\text{des}\perp}
    \label{total_ee_velocity}
\end{equation}\par
Similar to~\eqref{new_error_tot}, the error is computed at each step along the trajectory.
Contrary to~\eqref{new_error_tot}, subsequent error computations incorporate force sensor feedback.
This feedback is utilized to estimate the surface normal, $\hat{\mathbf{n}}_{\text{surf}}$, and to derive the projection matrices accordingly, thereby compensating for the surface friction bias.
Following the above explanations, the configuration space velocity command for the hybrid force-motion control can be expressed as follows:
\begin{equation}
      \mathbf{\dot{\Phi}}_{\text{cmd}}={\mathbf{J}_{\text{ee}}^\dagger}( \mathbf{{{v}}}_{\text{cmd}} +  \boldsymbol{{\xi}}_{h}) + \left(\mathbf{I} -  {\mathbf{J}_{\text{ee}}^\dagger}  \mathbf{J}_{\text{ee}}\right) \boldsymbol{\rho}
      \label{final_hybrid_command}
\end{equation}
Here, \(\boldsymbol{{\xi}}_{h}\) is the precision control component of the hybrid force-motion control command and is defined as follows:
\begin{equation}
    \boldsymbol{{\xi}}_{h} = \boldsymbol{{\xi}}_{m} +\boldsymbol{{\xi}}_{f}, \quad \boldsymbol{{\xi}}_{h} \in \mathbb{R}^{3 \times 1}
    \label{hybrid_error_}
\end{equation}
Substituting~\eqref{xi_1} and \eqref{xi_3} into~\eqref{hybrid_error_},
\begin{equation}
    \boldsymbol{{\xi}}_{h} = \mathbf{K}_{m \perp} \hspace{1pt}{\mathbf{e}}_{m} + \mathbf{K}_{f \perp} \hspace{1pt}{\mathbf{e}}_{f}
    \label{hybrid_error_definition}
\end{equation}
The equation~\eqref{final_hybrid_command} can be implemented to direct the robot to follow a desired velocity, denoted as \(\mathbf{{{v}}}_{\text{cmd}}\), for the end-effector \text{\emph{ee}}. 
The precision of this trajectory in motion-force directions is regulated by the gains, as defined in~\eqref{hybrid_error_definition}.\par
\subsection{Estimation of Surface Normal Using Online-Estimated Friction}
\begin{figure}[h]
    \centering
    \includegraphics[width=\linewidth]{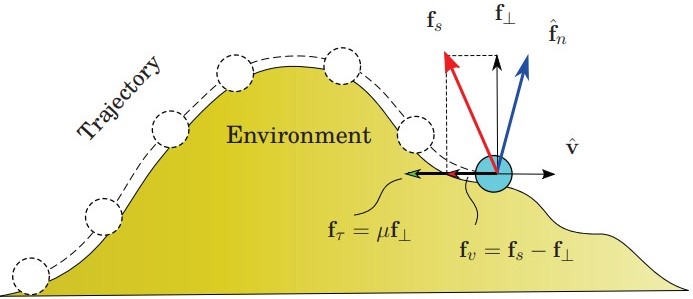}
    \caption{Estimation of Surface Normal Force Compensating for Environmental Friction.}
    \label{fig:friction_compensation}
\end{figure}

The novel method for surface normal estimation presented in this paper is a key component of the proposed hybrid force-motion framework. 
It dynamically updates the surface normal estimation in real-time, initially by constructing a surface friction model. 
This method primarily utilizes force sensing measurements and velocity commands to effectively compensate for friction bias.
In doing so, it ensures that the force controller can utilize the most current estimates of the surface normal as well as the Coulomb coefficient for enhanced control accuracy.
The algorithm adjusts the surface normal force, $\hat{\mathbf{f}_n}$, during the trajectory, as schematically shown in Fig.\ref{fig:friction_compensation}.\par
Algorithm.\ref{table2}, shows the surface normal estimation method.
In this approach, a feed-forward Coulomb friction force, denoted as \( \mathbf{f}_{\tau} \), is computed.
This computation relies on a projection force \( \mathbf{f}_{\perp} \) and \( \bar{\mu} \), where the latter represents a weighted moving average estimate of the friction coefficient.
The force \( \mathbf{f}_{\perp} \) is defined as the projection of the sensed force \(\mathbf{f}_s \), onto the null space of the robot's velocity, which is represented by \( \hat{\mathbf{v}} \). 
Consequently, the corrected surface normal force \( \hat{\mathbf{f}}_n \), is derived by subtracting \( \mathbf{f}_{\tau} \) from \( \mathbf{f}_s \).
The algorithm is executed at a frequency of 1 kHz. \\
In addition, at each time step, particularly at the \( k^{th} \) step, an updated measurement \( \mu_{k} \) is calculated using the two orthogonal components, \( \mathbf{f}_{\perp} \) and \( \mathbf{f}_v \). 
This measurement is subsequently integrated into the weighted moving average filter.
The estimated surface normal direction is finally determined by normalizing \( \hat{\mathbf{f}}_n \). 

\begin{algorithm*}[h]
\centering
\caption{: The Proposed Method for Enhanced Surface Normal Estimation with Friction Compensation}
\renewcommand{\arraystretch}{1.40} 
\begin{tabularx}{\textwidth}{lX}
\multicolumn{2}{c}{} \\
\hspace{6pt}\underline{\textbf{Given:}} & \textbf{"Description"}  \\
\hline %
$\mathbf{f}_s \in \mathbb{R}^3$ & Force sensor reading \\
\hdashline %
$\hat{\mathbf{v}} \in \mathbb{R}^3$ & Estimated velocity based on previous trajectory \\
\hdashline %
$\hat{\mathbf{M}}_a = [\hat{\mu}_{k-1}, \dots, \hat{\mu}_{k-m}]^\mathrm{T} $ & Previous \( m \) estimations of friction coefficient \\
\hdashline %
$\mathbf{w} = [w_1, \dots, w_m]^\mathrm{T} $ & Moving average weights \\
\hline %
\hspace{6pt} \underline{\textbf{Compute}}\hspace{3pt} "\( \hat{\mathbf{f}}_n": \) & \tikz \fill[pattern=north east lines] (0,0) rectangle (\linewidth,0.5cm); \\
\hline %
{\textbf{\textit{Step}} 1}:\hspace{6pt}  \textbf{if} \( \lVert \hat{\mathbf{v}} \rVert > v_{\epsilon} \) \textbf{then} & If the robot is moving, \( v_{\epsilon} \) - velocity threshold \\
\hdashline %
{\textbf{\textit{Step}} 2}: \hspace{12pt} \( \bar{\mathbf{\mu}} = \frac{1}{m} \sum_{i=1}^{m} w_i \hat{\mu}_{k-i} \)  & Weighted moving average on \( \hat{\mathbf{M}}_a \)  \\
\hdashline %
{\textbf{\textit{Step}} 3}: \hspace{12pt} \( \mathbf{\Omega}_{{\mathbf{v}}} = \hat{\mathbf{v}}(\hat{\mathbf{v}}^\mathrm{T} \hat{\mathbf{v}})^{-1} \hat{\mathbf{v}}^\mathrm{T},\hspace{5pt}  \mathbf{f}_{\perp} = (\mathbf{I} - \mathbf{\Omega}_{{\mathbf{v}}})\mathbf{f}_s
 \) & Projection onto "null space" of \( \hat{\mathbf{v}} \) \\
\hdashline %
{\textbf{\textit{Step}} 4}: \hspace{12pt} \( \mathbf{f}_{\tau} = -\bar{\mu} \lVert \mathbf{f}_{\perp} \rVert \frac{\hat{\mathbf{v}}}{\lVert \hat{\mathbf{v}} \rVert} \) & Compute surface friction \\
\hdashline %
{\textbf{\textit{Step}} 5}: \hspace{12pt} \( \hat{\mathbf{f}}_n = \mathbf{f}_s - \mathbf{f}_{\tau} \) & Obtain "surface normal force" by subtracting friction from force sensor reading \\
\hdashline %
{\textbf{\textit{Step}} 6}:\hspace{6pt}  \textbf{else} \( \hat{\mathbf{f}}_n = \mathbf{f}_s \) & \\
\hdashline %
{\textbf{\textit{Step}} 7}: \hspace{3pt} \textbf{end if} & \\
\hline %
\hspace{8pt}\underline{\textbf{Update}}\hspace{3pt} "\( \mu_k": \) & \tikz \fill[pattern=north east lines] (0,0) rectangle (\linewidth,0.5cm); \\
\hline %
{\textbf{\textit{Step}} 8}: \hspace{6pt} \textbf{if} \( \lVert \hat{\mathbf{v}} \rVert > v_{\epsilon} \) \textbf{then} & If the robot is moving, \( v_{\epsilon} \) - velocity threshold\\
\hdashline %
{\textbf{\textit{Step}} 9}: \hspace{13pt} \( \mathbf{f}_{v} = \mathbf{\Omega}_{{\mathbf{v}}}\mathbf{f}_s \) & Projection onto \( \hat{\mathbf{v}} \) direction \\
\hdashline %
{\textbf{\textit{Step}} 10}: \hspace{9pt} \( \mathbf{\mu}_k = \frac{\lVert \mathbf{f}_{v} \rVert}{\lVert \mathbf{f}_{\perp} \rVert} \) & Update current friction coefficient estimation \\
\hdashline %
{\textbf{\textit{Step}} 11}: \hspace{1pt} \textbf{else} \( \mathbf{\mu}_k = \mu_{k-1} \) & \\
\hdashline %
{\textbf{\textit{Step}} 12}: \hspace{1pt} \textbf{end if} & \\
\hline %
\multicolumn{2}{l}{\hspace{6pt}\underline{\textbf{Output:}}  \hspace{30pt} " \(\boldsymbol{\hat{\mathbf{f}}}_n , \hspace{3pt} \boldsymbol{\mu}_k\)"}
 \vspace{+6pt}\\
\label{table2}
\end{tabularx}
\vspace{-25pt} 
\end{algorithm*}

Following the above descriptions, along the trajectory, the F/T sensor reads the force data. An estimated velocity \(\hat{\mathbf{v}}\), and the friction coefficient \(\bar{\mu}\), are then considered, based on the data from the previous step. 
Subsequently, the force sensor reading \(\mathbf{f}_s \) is projected onto the null space of \(\hat{\mathbf{v}}\) using the following projection matrix.
\begin{equation}
    \mathbf{\Omega}_{{\mathbf{v}}} = \hat{\mathbf{v}}(\hat{\mathbf{v}}^\mathrm{T} \hat{\mathbf{v}})^{-1} \hat{\mathbf{v}}^\mathrm{T}
    \label{null-space_V}
\end{equation}
According to Fig.\ref{fig:friction_compensation}, projection of sensed force $\mathbf{f}_s$ onto robot velocity is given by,
\begin{equation}
    \mathbf{f}_{v} =\mathbf{f}_s - \mathbf{f}_{\perp}
\end{equation}
It is understood that \(\mathbf{f}_v\) represents the projection of the sensed force onto the direction of the robot's velocity.
    \begin{equation}
    \mathbf{f}_{v} = \mathbf{\Omega}_{{\mathbf{v}}}\mathbf{f}_s
    \label{force_sensor_projection}
\end{equation}
Thus, the force sensor projection onto the null space of the robot's velocity can be given as follows:
\begin{equation}
    \mathbf{f}_{\perp} = (\mathbf{I} - \mathbf{\Omega}_{{\mathbf{v}}})\mathbf{f}_s
    \label{projected fs}
\end{equation}
Then, a weighted moving average \(\bar{\mu}\), is calculated as follows:
\begin{equation}
    \bar{\mu} = \frac{1}{m} \sum_{i=1}^{m} w_i \hat{\mu}_{k-i}
    \label{average_mu}
\end{equation}
Here, \({w}_{i}\) represents a weight within the set of moving average weights:
\begin{equation}
    \mathbf{w} = [w_1, \dots, w_m]^\mathrm{T}
\end{equation}
Given~\eqref{projected fs},\eqref{average_mu} and understanding that the general surface friction is defined as follows:
\begin{equation}
    \mathbf{f}_{\tau}=\mu \mathbf{f}_{\perp}
\end{equation}
We can accordingly recompute the surface friction:
\begin{equation}
    \mathbf{f}_{\tau} = -\bar{\mu} \lVert \mathbf{f}_{\perp} \rVert \frac{\hat{\mathbf{v}}}{\lVert \hat{\mathbf{v}} \rVert}
    \label{computed_friction}
\end{equation}
Referring to Fig.\ref{fig:friction_compensation}, the surface normal force can be computed by subtracting the force sensor reading from the computed friction force, 
\begin{equation}
    \hat{\mathbf{f}}_n = \mathbf{f}_s - \mathbf{f}_{\tau}
    \label{normal_force}
\end{equation}
By normalizing this estimated surface normal force, we can estimate the direction of the surface normal at each step of the robot's trajectory.
\begin{equation} 
  \hat{\mathbf{n}}_{\text{surf}} = \frac{\hat{\mathbf{f}}_n}{\|\hat{\mathbf{f}}_n\|}
  \label{surface_normal}
\end{equation}
\begin{figure}
\centering
\includegraphics[width=1\columnwidth]{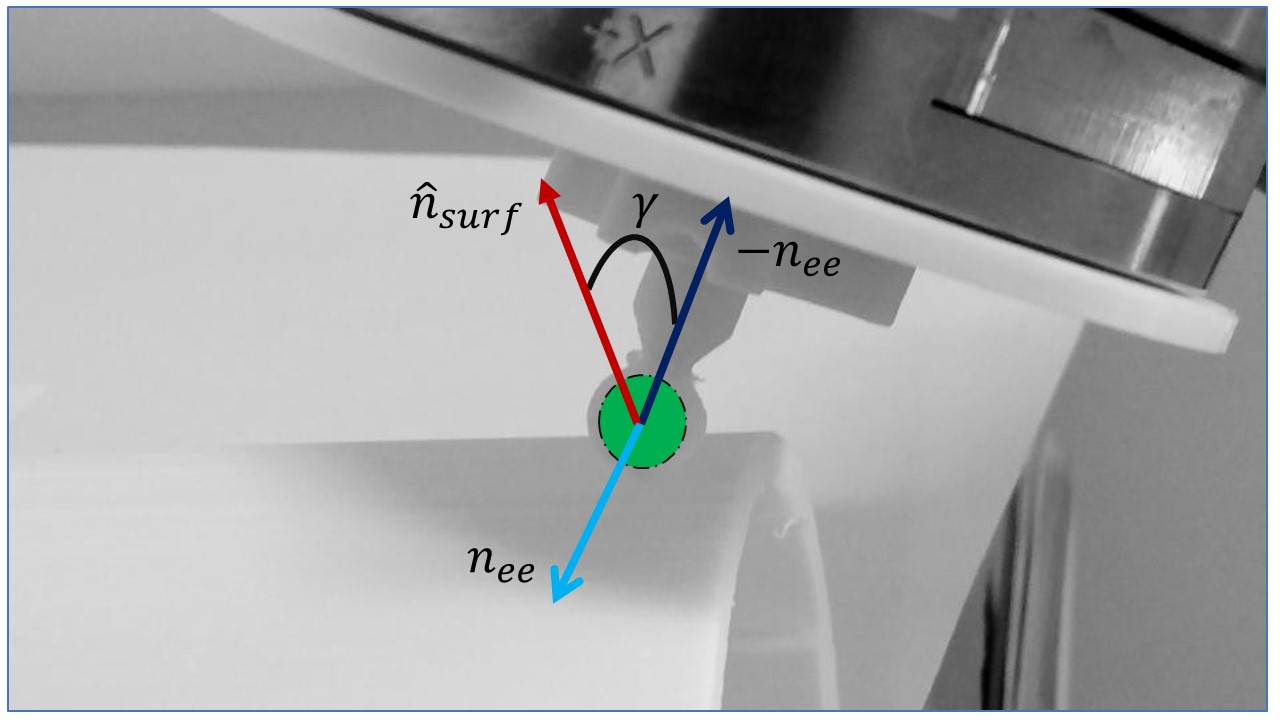}
\caption{Optimizing Probe Orientation: Utilizing Estimated Surface Normal and End-Effector Directions.}
\label{surface_norm}
\end{figure}
Furthermore, the surface friction coefficient, as detailed in~\eqref{computed_friction} , is continuously updated at each step of the trajectory.
This process begins with the projection of the force sensor reading \(\mathbf{f}_s\) onto the direction of the velocity \(\hat{\mathbf{v}}\), as described in~\eqref{force_sensor_projection}.
Subsequently, by utilizing the projection of the force sensor reading data onto both the direction of \(\hat{\mathbf{v}}\) and its null space, the friction coefficient is calculated as follows:
\begin{equation}
    \mu_k = \frac{\lVert \mathbf{f}_{v} \rVert}{\lVert \mathbf{f}_{\perp} \rVert}
\end{equation}

\subsection{Optimizing Probe Orientation}
In~\eqref{final_hybrid_command}, \(\boldsymbol{\rho}\) can be chosen to maximize the objective function $\mathbf{g}(\Phi)$. 
This function aims to optimize the alignment of the end-effector, \text{\emph{ee}}, with the estimated surface normal $\mathbf{\hat{n}}_{\text{surf}}$, by minimizing the angle \(\boldsymbol{\gamma}\), as depicted in Fig.\ref{surface_norm}. The formulation of this optimization problem is as follows: 
\begin{figure}
    \centering
    \includegraphics[width=1\columnwidth]{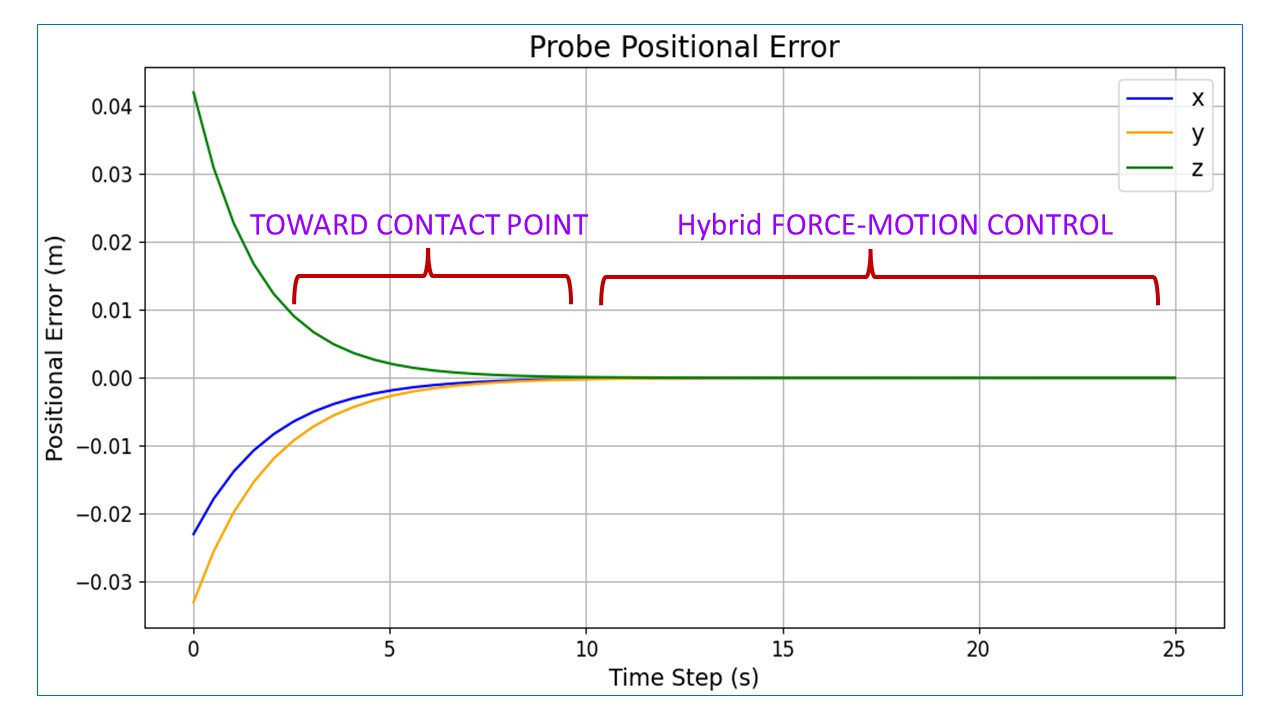}
    \caption{Tracking Hybrid Motion-Force Errors of the Probe.}
    \label{error}
\end{figure}
\begin{align}
   & \boldsymbol{\rho}=\alpha \;\nabla_{\Phi}{g},   \quad   \alpha > 0, \hspace{2pt} \alpha \in \mathbb{R}\label{objective}  \\[3pt]
   & {g}(\Phi)= \cos \gamma =  \mathbf{\hat{n}}_{\text{surf}}^\mathrm{T} \hspace{2pt}(-{\mathbf{n}}_{\text{ee}}) \label{gradien}
\end{align}
Here, $\alpha$ is a positive scalar, and the following relationship is then established:
\begin{equation}
    \begin{aligned}
        \underset{g}{\text{maximize}} \quad & \mathbf{\hat{n}}_{\text{surf}}^\mathrm{T} \hspace{2pt}(-{\mathbf{n}}_{\text{ee}}) \\
        \text{subject to} \quad & \mathbf{v} = \mathbf{v}_{\text{cmd}}
    \end{aligned}
    \label{optimization}
\end{equation} 
Here, \(\mathbf{\hat{n}}_{\text{surf}}\) denotes the estimated direction of the surface normal. This is derived using the algorithm proposed in the previous section.
\({\mathbf{n}}_{\text{ee}}\), on the other hand, represents the direction of the end-effector \emph{ee}, and can be derived from the robot's forward kinematics.
To explain how equation~\eqref{objective} optimizes the probe orientation, it is necessary to define $\nabla_{\Phi}{g}$. 
To do so, we take the derivative of the objective function \emph{g} as follows:
 \begin{align}
    &\dot{g} = \underbrace{\cancel{-\dot{\hat{\mathbf{n}}}_{\text{surf}}^\mathrm{T} \mathbf{n}_{\text{ee}}}}_{\xrightarrow{\hspace{12pt}}0} - {\hat{\mathbf{n}}}_{\text{surf}}^\mathrm{T} \dot{\mathbf{n}}_{\text{ee}} \\
    &\textit{Thus},\ \dot{g} = - {\hat{\mathbf{n}}}_{\text{surf}}^\mathrm{T} \hspace{1pt} \dot{\mathbf{n}}_{\text{ee}} \label{gradient_g}
\end{align}
Here, $\dot{\mathbf{n}}_{\text{ee}}$ represents the velocity of the end-effector, and it can also be given by,
 \begin{equation}
     \dot{\mathbf{n}}_{\text{ee}} = \boldsymbol{\omega} \times \mathbf{n}_{\text{ee}}, \quad \boldsymbol{\omega} \in \mathbb{R}^{3 \times 1} \label{angular_v_endeffector}
 \end{equation}
Here, $\boldsymbol{\omega}$ represents the angular velocity of the \textit{ee}. By substituting~\eqref{angular_v_endeffector} into~\eqref{gradient_g} and utilizing the properties of the vector cross product:
\begin{equation}
    \dot{g}=  - {\hat{\mathbf{n}}}_{\text{surf}}^\mathrm{T} \hspace{2pt} \underbrace{\big{(}[\mathbf{n}_{\text{ee}}]_{\times}^{\mathrm{T}} \hspace{2pt} \boldsymbol{\omega} \big{)}}_{\triangleq \hspace{1pt}\dot{\mathbf{n}}_{\text{ee}}} \label{skew_cross_product}
\end{equation}
Where, $[\mathbf{n}_{\text{ee}}]_\times$ represents the Skew-symmetric matrix operator:
\begin{equation}
    [\mathbf{n}_{\text{ee}}]_\times = 
    \begin{bmatrix}
        0 & -\mathbf{n}_{\text{ee,3}} & \mathbf{n}_{\text{ee,2}} \\
        \mathbf{n}_{\text{ee,3}} & 0 & -\mathbf{n}_{\text{ee,1}} \\
        -\mathbf{n}_{\text{ee,2}} & \mathbf{n}_{\text{ee,1}} & 0
    \end{bmatrix}
\end{equation}
Knowing that the angular velocity of (\textit{ee}), $\boldsymbol{\omega}$, is given by,
\begin{equation}
    \boldsymbol{\omega}=\mathbf{J}_{\text{ee,w}} \dot{\Phi}
\end{equation}
Here, $\mathbf{J}_{\text{ee,w}}$ represents the angular velocity component of the robot's end-effector Jacobian.
Equation~\eqref{skew_cross_product} can be rewritten:
\begin{align}
    \dot{g} \triangleq (\nabla_{\Phi}{g})^\mathrm{T} \hspace{2pt}\dot{\Phi}=  
    \underbrace{- {\hat{\mathbf{n}}}_{\text{surf}}^\mathrm{T} \hspace{2pt} [\mathbf{n}_{\text{ee}}]_{\times}^{\mathrm{T}} \hspace{2pt} \mathbf{J}_{\text{ee},\omega}}_{\triangleq \left(\nabla_{\Phi}{g}\right)^\mathrm{T}}
    \hspace{1pt}\dot{\Phi}
    \label{final_g_dot}
\end{align}
Considering~\eqref{objective},\eqref{gradien} and~\eqref{final_g_dot}, the definition of $\nabla_{\Phi}{g}$ is as follows:
\begin{align}
     \nabla_{\Phi}{g} = - 
     \left(\mathbf{J}_{\text{ee},\omega}\right)^\mathrm{T} \hspace{1pt}
     [\mathbf{n}_{\text{ee}}]_{\times}  \hspace{2pt}
     {\hat{\mathbf{n}}}_{\text{surf}} \hspace{2pt}
\end{align}

\section{Simulation and Experimental Validation}
\begin{figure}[t]
    \centering
    \includegraphics[width=\columnwidth]{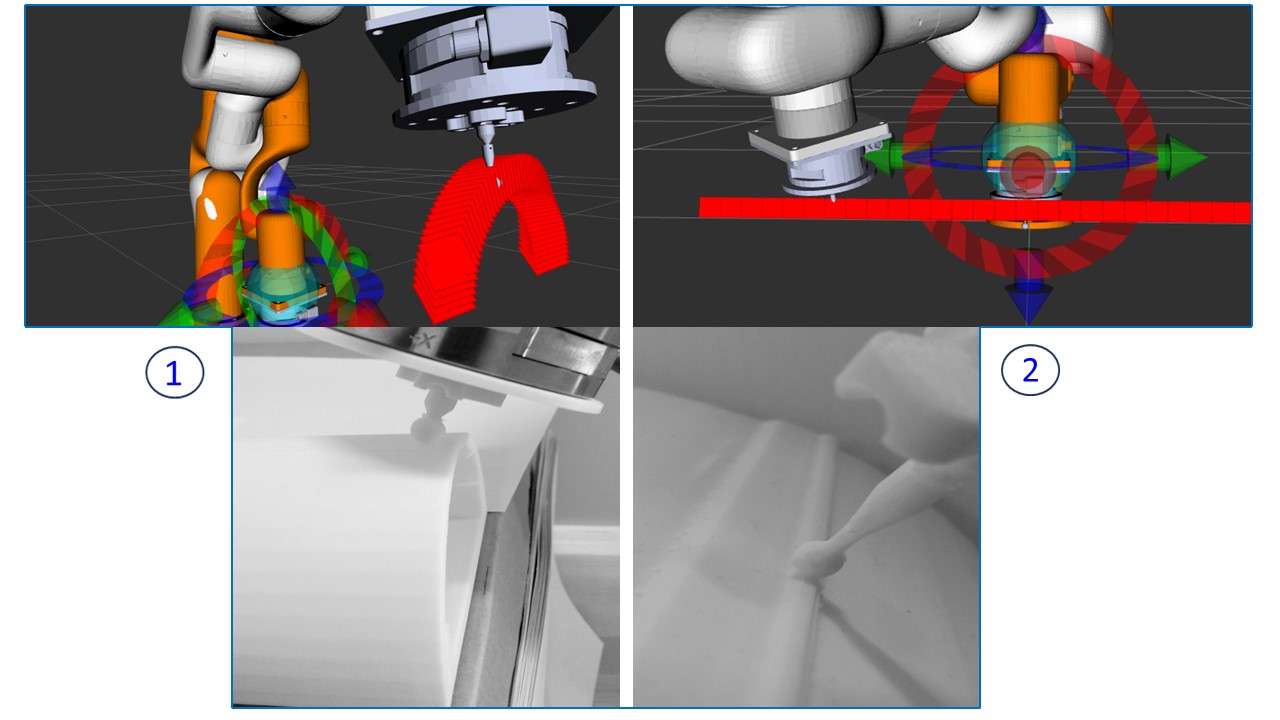}
    \caption{ ROS2 Simulation and Experimentation: 1) Estimating Surface Normal on a Dome-Shaped Workpiece, 2) Estimation of Surface Normal Along a Linear Path on a Workpiece.}
    \label{paths}
\end{figure}
Building upon the explanations provided in Section~\ref{system_modeling_section}, the robot moves towards the targeted contact point by utilizing a resolved rate algorithm.
Once the robot makes contact with the surface at the designated contact point, the hybrid force-motion controller initiates operation, using the command specified in~\eqref{final_hybrid_command}. The following gains, desired force, and offset have been selected for the controller:
\begin{align}
	& \mathbf{K}_{m}, \mathbf{K}_{f}= \text{diag}([10, 10, 10]), \\
	& \mathbf{K}_{\text{adm}} = \text{diag}([0.1, 0.1, 0.1])\\
        & \mathbf{f}_{\text{des}}= \begin{bmatrix} 0 & 0 & -2 \end{bmatrix}^{\mathrm{T}} \label{desired_force} \\
        & \mathbf{d}_h =\begin{bmatrix} 0 & 0 & 0.05 \end{bmatrix}^{\mathrm{T}} 
       \label{defined_gains}
\end{align}
Fig.\ref{error} illustrates an example of error convergence for the end-effector, as detailed in~\eqref{motion_error}, following a linear trajectory as depicted in Fig.\ref{paths}-2.
This indicates that the probe accurately reached the contact point on the workpiece.
It then begins following the trajectory, using the hybrid force-motion command from~\eqref{final_hybrid_command}, starting around time step 10 and continuing to the end. 
This process updates the surface normal in accordance with the method outlined in Algorithm\ref{table2}.\par 
To assess the accuracy of the proposed approach, a 7-DoF manipulator equipped with a probe at its end-effector is considered. An F/T sensor is integrated at the wrist of the manipulator, as illustrated in Fig\ref{system}.
During the testing phase, which included both simulations and experiments, a diverse range of trajectories on workpieces were assessed. 
These included linear, sinusoidal, and curved paths, as well as paths resembling a semi-circular arc(dome), among others.
The goal was to evaluate the precision of the surface normal estimation method within the hybrid force-motion control framework. This evaluation focused on two scenarios: one where the surface normal direction remains constant, and another where it varies, such as when following paths along an arc-shaped contour.\par
Fig.\ref{paths} showcases some of these path trajectories, illustrated both in a ROS2 simulation environment and in their corresponding experimental setups.
The trajectories are categorized as follows:
\begin{itemize}
    \item  A trajectory where the direction of the surface normal remains unchanged.
    \item  A trajectory where the direction of the surface normal may change.
\end{itemize}
\begin{figure}[t]
    \centering
    \begin{subfigure}{\columnwidth}
        \centering
        \includegraphics[width=0.95\linewidth]{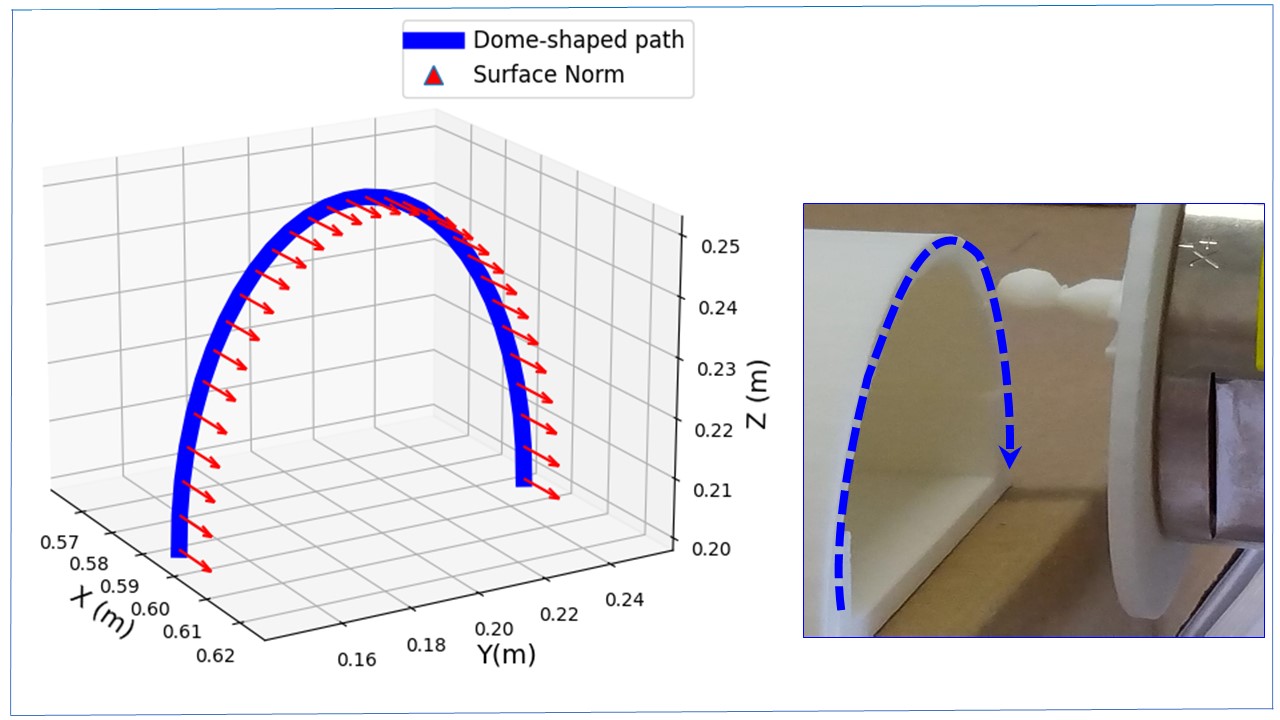}
        \caption{Surface normal in the X-axis direction remains unchanged throughout the trajectory}
        \label{subfig:a}
    \end{subfigure}
    \begin{subfigure}{\columnwidth}
        \centering
        \includegraphics[width=0.95\linewidth]{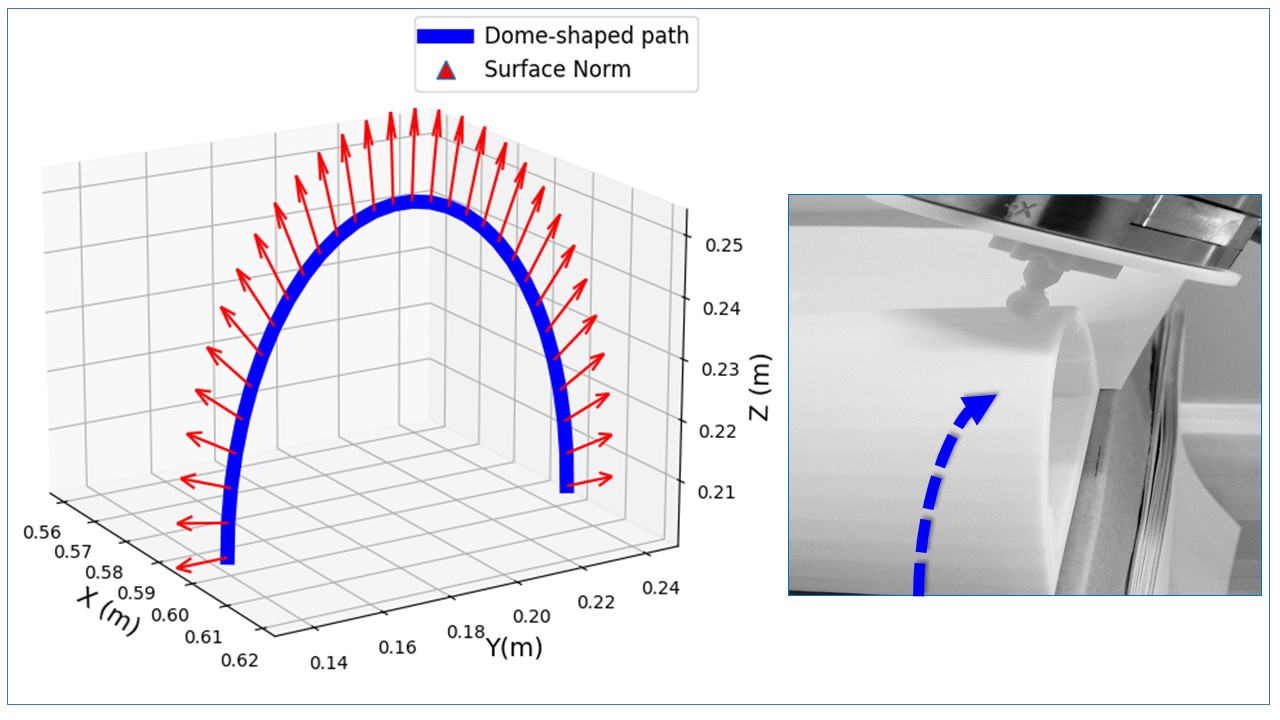}
        \caption{Variation in surface normal direction during the trajectory}
        \label{subfig:b}
    \end{subfigure}
    \caption{Surface Normal Plots for Various Trajectories: (a) Surface Normal with a Fixed Direction, and (b) Surface Normal that Changes Throughout the Trajectory.}
    \label{Norm_plots}
\end{figure}
\begin{figure}
    \centering
    \begin{subfigure}{\columnwidth}
        \centering
        \includegraphics[width=\linewidth]{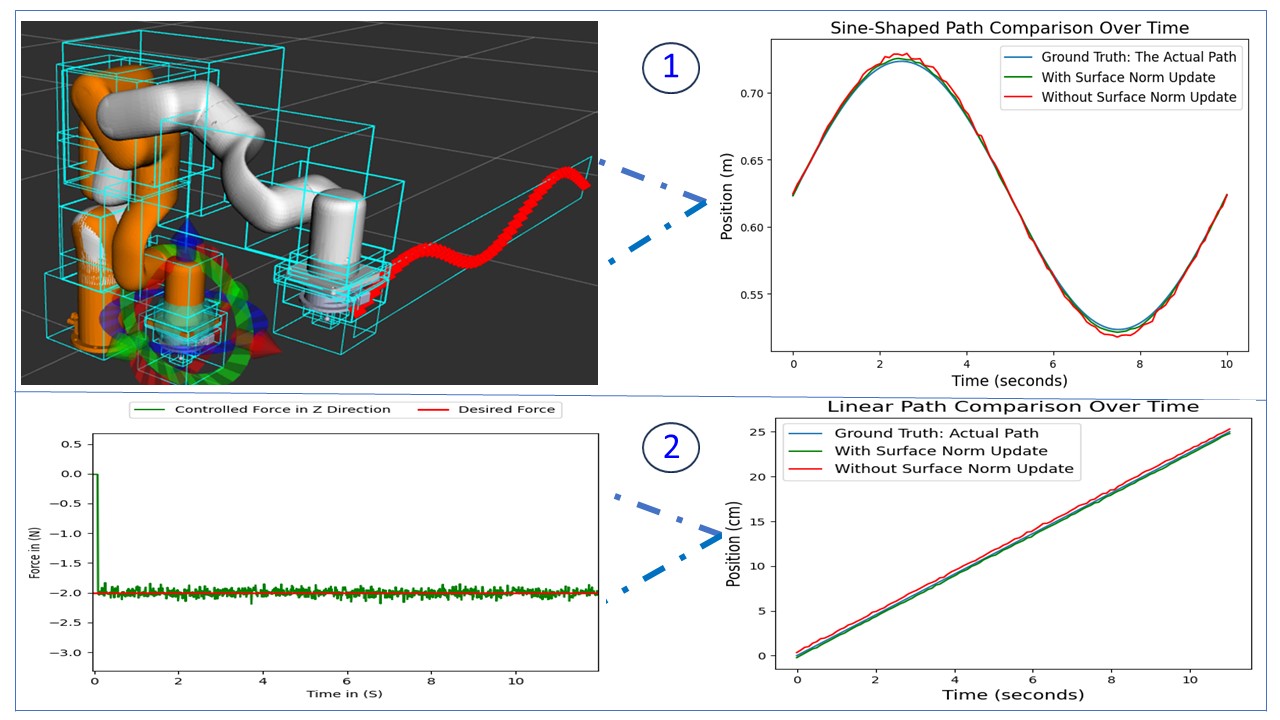}
        \caption{ Hybrid force-motion control, with and without surface normal estimation method on two different paths: 1) A sine-shaped path, and 2) A linear path with it's force control plot}
        \label{subfig_10_a}
    \end{subfigure}
    \hfill
    \begin{subfigure}{\columnwidth}
        \centering
        \includegraphics[width=\linewidth]{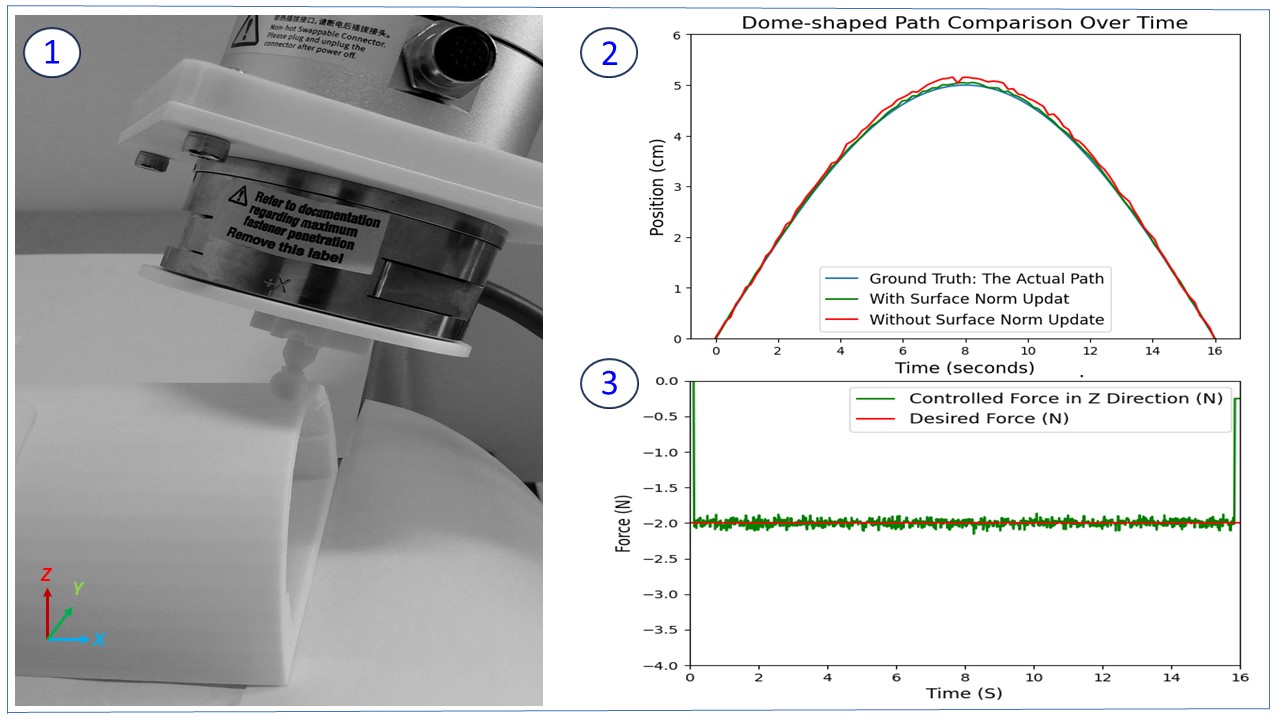}
        \caption{Tracking error for a dome-shaped workpiece: 1- Probe Path following on the workpiece in experimental Setup, 2-  Force control error, with and without surface normal update for the dome-shaped object, and 3- Force control plot.}
        \label{subfig_b}
    \end{subfigure}
    \caption{Plots of Position and Force Tracking Errors for: a) A Sine-Shaped and a Linear Paths, and  b) A Dome-Shaped Path.}
    \label{error_plots}
\end{figure}
In the first image of Fig.\ref{Norm_plots}, the trajectory of the robot's end-effector (\text{ee}) for the hybrid force-motion task on the dome-shaped workpiece is in Y-Z plane.
Throughout the path trajectory, as indicated by the blue dashed line, the estimated surface normals are parallel to the X-axis.
Conversely, in the second image of Fig.\ref{Norm_plots}, the robot traverses over a dome-shaped object. 
Here, the estimated surface normals have different directions along the path.\par
The surface normal is calculated and continuously updated based on feedback from the force sensor, which is leveraged for surface friction bias compensation, as shown in~\eqref{surface_normal}.
Therefore, it's essential to evaluate the accuracy of this method, especially in tracking positional-force errors on surfaces of differently shaped objects. 
This evaluation will ascertain whether the proposed normal estimation method enables more accurate tracing compared to methods with non-updated normals.\par
In Fig.\ref{error_plots}, the red plots represent the probe's (\text{\emph{ee}}) hybrid force-motion trajectory over variously shaped surfaces, conducted without using the algorithm proposed in this paper. 
Conversely, the green plots illustrate the results obtained using the proposed surface normal estimation algorithm, as detailed in Algorithm\ref{table2}.
It is evident that the positional tracking error on the surfaces is reduced with the proposed method across all differently shaped surfaces.
The improved accuracy in these plots is particularly noticeable on curved shapes, especially at the peaks or troughs.
In the dome-shaped and sinusoidal-shaped paths, the differences between the target and actual paths are 0.3mm at minimum and 1mm at maximum, respectively.
The error becomes smaller in paths with an unchanged surface normal, as shown in Fig.\ref{subfig:a} and \ref{subfig_10_a}-2, reducing to approximately 0.3mm and 0.8mm, respectively.
The average accuracy increase achieved by the proposed method is approximately 5\%, compared to the method without surface norm updates. In addition, in both Fig.\ref{subfig_10_a} and Fig.\ref{subfig_b}, the end-effector (\textit{ee}) controlled force, followed precisely with the desired force, as detailed in~\eqref{desired_force}. \par
\vspace{5pt}
\section{Conclusion}
In this paper, we introduce a novel method for real-time surface normal estimation.
It leverages force sensor readings and velocity commands, compensating for surface friction within a hybrid force-motion control framework.
The proposed algorithm is implemented in a \text{\emph{ROS2}} environment and has been tested through both simulation and experiment on multiple workpieces with variously shaped surfaces. 
These include a line, a sine-shaped path, a dome-shaped, among others.
The current implementation delivers satisfactory performance, as illustrated in the accompanying figures and plots.
Utilizing the estimated surface normal method, the position-force errors tracking of the explored paths — especially on surfaces where the normal changes direction along the path — aligns very closely with the actual path.
The proposed method has increased positional tracking accuracy by up to 5\%, compared to previous methods where the surface normal is not updated during the trajectory.\par

\vspace{12pt}


\begin{thebibliography}{00}


\bibitem{Stolt2012} {A. Stolt, M. Linderoth, A. Robertsson and R. Johansson, "Force controlled robotic assembly without a force sensor," 2012 IEEE International Conference on Robotics and Automation, Saint Paul, MN, USA, 2012, pp. 1538-1543, doi: 10.1109/ICRA.2012.6224837.}
\bibitem{Yin2012}{Y. H. Yin, Y. Xu, Z. H. Jiang and Q. R. Wang, "Tracking and Understanding Unknown Surface With High Speed by Force Sensing and Control for Robot," in IEEE Sensors Journal, vol. 12, no. 9, pp. 2910-2916, Sept. 2012, doi: 10.1109/JSEN.2012.2205098.}

\bibitem{Thomessen2000} {Thomessen, Trygve and Lien, Terje. (2000). Robot control system for safe and rapid programming of grinding applications. Industrial Robot: An International Journal. 27. 437-444. 10.1108/01439910010355766. }


\bibitem{Ahrens1998} {M. Ahrens, V. Mallick and K. Parfrey, "Robotic based thermoplastic fibre placement process," Proceedings. 1998 IEEE International Conference on Robotics and Automation (Cat. No.98CH36146), Leuven, Belgium, 1998, pp. 1148-1153 vol.2, doi: 10.1109/ROBOT.1998.677247.}

\bibitem{Ohba2009} {Y. Ohba et al., "Sensorless Force Control for Injection Molding Machine Using Reaction Torque Observer Considering Torsion Phenomenon," in IEEE Transactions on Industrial Electronics, vol. 56, no. 8, pp. 2955-2960, Aug. 2009, doi: 10.1109/TIE.2009.2024444.}

\bibitem{Huang2009}{H. B. Huang, D. Sun, J. K. Mills and S. H. Cheng, "Robotic Cell Injection System With Position and Force Control: Toward Automatic Batch Biomanipulation," in IEEE Transactions on Robotics, vol. 25, no. 3, pp. 727-737, June 2009, doi: 10.1109/TRO.2009.2017109.}

\bibitem{Ferretti1997} {G. Ferretti, G. Magnani and P. Rocco, "Toward the implementation of hybrid position/force control in industrial robots," in IEEE Transactions on Robotics and Automation, vol. 13, no. 6, pp. 838-845, Dec. 1997, doi: 10.1109/70.650162.}

\bibitem{Natale1999} {C. Natale, B. Siciliano and L. Villani, "Robust hybrid force/position control with experiments on an industrial robot," 1999 IEEE/ASME International Conference on Advanced Intelligent Mechatronics (Cat. No.99TH8399), Atlanta, GA, USA, 1999, pp. 956-960, doi: 10.1109/AIM.1999.803301.}
\bibitem{Yoshikawa1993}{T. Yoshikawa and A. Sudou, "Dynamic hybrid position/force control of robot manipulators-on-line estimation of unknown constraint," in IEEE Transactions on Robotics and Automation, vol. 9, no. 2, pp. 220-226, April 1993, doi: 10.1109/70.238286.}

\bibitem{Solanes2018} {Solanes, J. E., Gracia, L., Muñoz-Benavent, P., Valls Miro, J., Perez-Vidal, C., and Tornero, J. (November 28, 2018). "Robust Hybrid Position-Force Control for Robotic Surface Polishing." ASME. J. Manuf. Sci. Eng. January 2019; 141(1): 011013. https://doi.org/10.1115/1.4041836}

\bibitem{Namvar2005} {M. Namvar and F. Aghili, "Adaptive force-motion control of coordinated robots interacting with geometrically unknown environments," in IEEE Transactions on Robotics, vol. 21, no. 4, pp. 678-694, Aug. 2005, doi: 10.1109/TRO.2004.842346.}

\bibitem{Xia2016} {G. Xia, C. Li, Q. Zhu and X. Xie, "Hybrid force/position control of industrial robotic manipulator based on Kalman filter," 2016 IEEE International Conference on Mechatronics and Automation, Harbin, China, 2016, pp. 2070-2075, doi: 10.1109/ICMA.2016.7558885.}

\bibitem{Wang2021}{Ziling Wang, Lai Zou, Xiaojie Su, Guoyue Luo, Rui Li, and Yun Huang. 2021. Hybrid force/position control in workspace of robotic manipulator in uncertain environments based on adaptive fuzzy control. Robot. Auton. Syst. 145, C (Nov 2021). https://doi.org/10.1016/j.robot.2021.103870}

\bibitem{Bona2005}{B. Bona and M. Indri, “Friction compensation in robotics: An overview,”in Proc. 44th IEEE Conf. Decision Control Eur. Control Conf., Seville,
Spain, Dec. 15–15, 2005, pp. 4360–4367.}

\bibitem{Dupont94} {P. E. Dupont, “Avoiding stick-slip through PD control,” IEEE Trans. Autom.
Control, vol. 39, no. 5, pp. 1094–1097, May 1994.}

\bibitem{Olsson1999}{H. Olsson, K. J. Astrm, C. Canudas de Wit, M. Gfvert, and P. Lischinsky,
“Friction models and friction compensation,” Eur. J. Control, vol. 4, no. 3,
pp. 176–195, 1998.}

\bibitem{Misovec1999} {K. M. Misovec and A. M. Annaswamy, “Friction compensation using
adaptive nonlinear control with persistent excitation,” Int. J. Control,
vol. 72, no. 5, pp. 457–479, 1999.}

\bibitem{Kermani2004} {M. Kermani, M. Wong, R. Patel, M. Moallem, and M. Ostojic, “Friction compensation in low and high-reversal-velocity manipulators,” in Proc.
IEEE Int. Conf. Robotics Automation, New Orleans, LA, USA, Apr. 26–May 1, 2004, pp. 4320–4325.}

\bibitem{Avelar2016} {C. Aguilar-Avelar and J. Moreno-Valenzuela, “New feedback linearization-based control for arm trajectory tracking of the Furuta Pendulum,” IEEE/ASME Trans. Mechatronics, vol. 21, no. 2, pp. 638–648, Apr. 2016.}

\bibitem{Huang2016} {S. Huang, W. Liang and K. K. Tan, "Intelligent Friction Compensation: A Review," in IEEE/ASME Transactions on Mechatronics, vol. 24, no. 4, pp. 1763-1774, Aug. 2019, doi: 10.1109/TMECH.2019.2916665.}


\bibitem{Jatta2006} {F. Jatta, G. Legnani and A. Visioli, "Friction compensation in hybrid force/velocity control of industrial manipulators," in IEEE Transactions on Industrial Electronics, vol. 53, no. 2, pp. 604-613, April 2006, doi: 10.1109/TIE.2006.870682.}
\bibitem{Cao2019} {H. Cao, X. Chen, Y. He and X. Zhao, "Dynamic Adaptive Hybrid Impedance Control for Dynamic Contact Force Tracking in Uncertain Environments," in IEEE Access, vol. 7, pp. 83162-83174, 2019, doi: 10.1109/ACCESS.2019.2924696.}


\bibitem{Hess1990} {D. P. Hess and A. Soom, “Friction at a lubricated line contact operating at oscillating sliding velocities,” Trans. ASME J. Tribology, vol. 112, no. 1,
pp. 147–152, 1990.}


\bibitem{Canudas1995} {C. Canudas DeWit, C. H. Olsson, K. J. Astrm, and P. Lischinsky, “A new model for control of systems with friction,” IEEE Trans. Autom. Control, vol. 40, no. 3, pp. 419–425, Mar. 1995.}

\bibitem{Swevers2000} {J. Swevers, F. Al-Bender, C. Ganseman, and T. Prajogo, “An integrated friction model structure with improved presliding behaviour for accurate
friction compensation,” IEEE Trans. Autom. Control, vol. 45, no. 4, pp. 675–686, Apr. 2000.}

\bibitem{Lampaert2002} {V. Lampaert, J. Swevers, and F. Al-Bender, “Modification of the Leuven integrated friction model structure,” IEEE Trans. Autom. Control, vol. 47,
no. 4, pp. 683–687, Apr. 2002.}



\bibitem{Chen2009} {S. L. Chen, K. K. Tan, and S. Huang, “Friction modeling and compensation of servo-mechanical system with dual-relay feedback approach,”
IEEE Trans. Control Syst. Technol., vol. 17, no. 6, pp. 1295–1305, Nov.2009.}

\bibitem{Lee1995} {S. W. Lee and J. H. Kim, “Robust adaptive stick-slip friction compensation,” IEEE Trans. Ind. Electron., vol. 42, no. 5, pp. 474–479, Oct.
1995.}

\bibitem{Yang2018} {X. Yang et al., "Continuous Friction Feedforward Sliding Mode Controller for a TriMule Hybrid Robot," in IEEE/ASME Transactions on Mechatronics, vol. 23, no. 4, pp. 1673-1683, Aug. 2018, doi: 10.1109/TMECH.2018.2853764.}



\bibitem{Qian2019}{Y. Qian, J. Yuan, S. Bao and L. Gao, "Sensorless Hybrid Normal-Force Controller With Surface Prediction," 2019 IEEE International Conference on Robotics and Biomimetics (ROBIO), Dali, China, 2019, pp. 83-88, doi: 10.1109/ROBIO49542.2019.8961532.}

\bibitem{Li2022} {Li, Jian, Guan, Yisheng, Chen, Haowen, Wang, Bing, Zhang, Tao, Hong, Jie and Wang, Danwei. (2022). Real-time normal contact force control for robotic surface processing of workpieces without a priori geometric model. The International Journal of Advanced Manufacturing Technology. 119. 10.1007/s00170-021-07497-2. }

\bibitem{Khatib1987} {Khatib, O., 1987, “A Unified Approach for Motion and Force Control of Robot Manipulators: The Operational Space Formulation,” IEEE J. Rob. Autom.,
RA-3(1), pp. 43–53.}

\bibitem{Featherstone1999} {Featherstone, R., Thiebaut, S., and Khatib, O., 1999, “A General Contact Model for Dynamically-Decoupled Force/Motion Control,” IEEE International Conference
on Robotics and Automation (ICRA), Detroit, MI, May 10–15, pp. 3281–3286.}

\bibitem{Whitney1969} {Whitney, D. E., 1969, “Resolved Motion Rate Control of Manipulators and Human Prostheses,” IEEE Trans. Man-Mach. Syst., 10(2), pp. 47–53.}

\end{thebibliography}
\end{document}